\documentclass[10pt,twocolumn,letterpaper]{article}

\usepackage{3dv}
\usepackage{times}
\usepackage{epsfig}
\usepackage{graphicx}
\usepackage{amsmath}
\usepackage{amssymb}

\usepackage{booktabs, multirow} %
\usepackage{changepage,threeparttable}
\usepackage{paralist}

\usepackage[pagebackref=true,breaklinks=true,letterpaper=true,colorlinks,bookmarks=false]{hyperref}

\threedvfinalcopy %

\def\threedvPaperID{234} %

\ifthreedvfinal\fi
\begin{document}

\title{Spatial Attention Improves Iterative 6D Object Pose Estimation}

\author{{Stefan Stev{\v{s}}i{\'{c}} \quad Otmar Hilliges}\\
Department of Computer Science, ETH Zürich, Switzerland\\
{\tt\small \{stefan.stevsic, otmar.hilliges\}@inf.ethz.ch}
}

\maketitle
\thispagestyle{empty}

\begin{abstract}
The task of estimating the 6D pose of an object from RGB images can be broken down into two main steps: an initial pose estimation step, followed by a refinement procedure to correctly register the object and its observation.
In this paper, we propose a new method for 6D pose estimation refinement from RGB images.
To achieve high accuracy of the final estimate, the observation and a rendered model need to be aligned.   
Our main insight is that after the initial pose estimate, it is important to pay attention to distinct spatial features of the object in order to improve the estimation accuracy during alignment.
Furthermore, parts of the object that are occluded in the image should be given less weight during the alignment process.
Most state-of-the-art refinement approaches do not allow for this fine-grained reasoning and can not fully leverage the structure of the problem.
In contrast, we propose a novel neural network architecture built around a spatial attention mechanism that identifies and leverages information about spatial details during pose refinement. We experimentally show that this approach learns to attend to salient spatial features and learns to ignore occluded parts of the object, leading to better pose estimation across datasets.
We conduct experiments on standard benchmark datasets for 6D pose estimation (LineMOD and Occlusion LineMOD) and outperform previous state-of-the-art methods.  
\end{abstract}

\section{Introduction}
\label{sec:intro}
Detecting objects in images and estimating their pose from 2D images is one of the core problems in computer vision and has many applications in downstream tasks. 
With the advent of deep learning approaches, rapid progress on the task of object detection has been achieved (e.g., YOLO \cite{redmon2016you}, Fast R-CNN \cite{girshick2015fast}, and Mask R-CNN \cite{he2017mask}). 
However, for applications such as robotic manipulation, augmented reality and autonomous driving, it is essential to recover the full 6D pose of the object. 
To solve this challenging task, a variety of methods such as BB8 \cite{rad2017bb8}, PoseCNN \cite{xiang2017posecnn}, PVNet \cite{peng2019pvnet} and DPOD \cite{zakharov2019dpod} have been proposed recently. The 6D pose estimation task is typically separated into two subtasks: initial detection and pose estimation of objects from raw images \cite{peng2019pvnet, hinterstoisser2011multimodal, hu2019segmentation, xiang2017posecnn, rad2017bb8} and subsequent pose refinement \cite{li2018deepim, zakharov2019dpod}. With modern CNN architectures, the one-shot pose estimation sub-task has saturated somewhat \cite{rad2017bb8, xiang2017posecnn, peng2019pvnet, zakharov2019dpod, hu2019segmentation} but such methods have not yet been able to precisely align spatial details, leaving much room for improvement via iterative pose refinement, which is the focus of our work.

Traditionally initial pose estimation and refinement, often via the iterative closest point (ICP) algorithm, have been treated separately.   
More recently, several methods \cite{li2018deepim, zakharov2019dpod} address the refinement problem using deep learning approaches, leading to superior performance compared to ICP refinement. 
The DPOD refiner \cite{zakharov2019dpod} leverages a pre-trained ResNet model \cite{he2016deep} and leverages it for 6D pose refinement in a non-iterative fashion. 
DeepIM \cite{li2018deepim} utilizes an iterative strategy to refine the prediction through several stages.
While achieving good accuracy neither of these methods explicitly leverages existing structure of the alignment process. For example, accurate alignment of 3D model and 2D image observation requires consideration of high-frequency spatial features. It has also been shown that excluding occlusions from the input data leads to more accurate predictions \cite{hu2019segmentation}. However, existing pose refinement approaches do not explicitly take this into consideration.

\begin{figure*}
    \centering
    \includegraphics[width = 0.98 \textwidth]{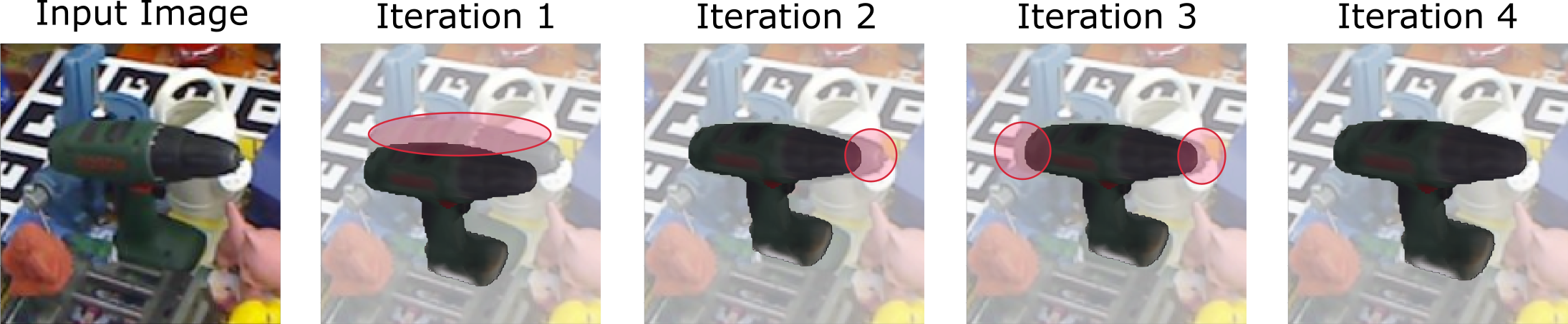}
    \caption{Iterative pose refinement. The goal of the refinement process is to align the model and the input image. At each iteration stage, we focus on aligning a different feature. At the first iteration, we align top edges; at the second, we align the front point of the driller; at the third, we scale the model to cover non-overlapping details. This results in perfect alignment.}
    \label{fig:teasr}
\end{figure*}

To address this issue we propose a novel deep neural network architecture that learns to consider fine-grained spatial details and learns to reason about occlusions in the scene. 
An important insight is that a small number of pixels are sufficient to align the model to the input image. Humans would perform this task  by first aligning some characteristic feature, for example, corners of the object. Then, geometric detail is leveraged to adjust for scale and the entire process would be repeated iteratively, by focusing on details that most reliably show the displacement (see Fig. \ref{fig:teasr}).
Inspired by this scheme, SoA methods for one-shot pose estimation rely on RANSAC to extract the most reliable points for prediction \cite{peng2019pvnet, zakharov2019dpod}. 
We also leverage the idea that a small number of specific features is important for alignment, but we use an attention mechanism to learn to identify these automatically. We introduce an attention model that isolates reliable features, while rejecting the clutter. This makes the entire pipeline differentiable and allows for application of the inlier features to direct pose refinement which would not be possible via RANSAC. Furthermore, a fully differentiable approach allows for fine-tuning of the pose estimator for a given downstream task.

When overlaying the image and the rendered prediction, similar points from real and rendered images lie nearby. Therefore, a spatial attention mechanism can be learned to identify the details which give the best position estimation if the final registration error is backpropagated through the entire network. As we experimentally show in Sec.~\ref{sec:att_model}, the attention mechanism focuses on object outlines, which contain a lot of discriminative details. Outline edges are a common choice for ICP-based refinement from RGB images. We use a soft attention mechanism, similar to \cite{jetley2018learn, kosiorek2017hierarchical}, which results in a fully differentiable approach that can be trained using standard loss functions. As a feature extracting backbone, we use U-shaped DenseNet from \cite{jegou2017one}. DenseNet produces good results even when trained on small datasets and its U-shaped variant shows excellent performance on the semantic segmentation task. This task requires correctly classifying small details, which is also useful in our task because the attention needs to detect particular details. We test our approach in the single-stage and iterative approach settings. The iterative approach enables the attention to specialize for specific details in each iteration. This leads to state-of-the-art performance on the standard datasets for 6D pose estimation. 

More specifically, we make the following contributions: 
\begin{inparaenum}[(i)]
    \item We introduce a new neural network model for 6D pose refinement that leverages an attention mechanism.
    \item We provide a detailed experimental evaluation of our model on standard datasets for 6D pose estimation in single-stage and multi-stage settings. We show that our model leads to state-of-the-art performance on these datasets.
    \item We experimentally show that attention mechanism enables the network to isolate the distinctive object details and exclude occlusions from the prediction. 
\end{inparaenum}

\section{Related Work}

The task of 6D pose estimation has been a long-standing problem in Computer Vision. Early methods \cite{hinterstoisser2011multimodal, hinterstoisser2012model, rios2013discriminatively} mostly rely on RGB-D images because it is possible to recover scale via depth information. The most popular dataset for evaluation of methods has been LineMOD \cite{hinterstoisser2011gradient}. However, the LineMOD dataset does not contain occluded test objects. When the object is occluded, it is much harder to distinguish between the object of interest and occlusions. Thus, the accuracy of estimation on occlusion data is far below the non-occluded case \cite{peng2019pvnet, zakharov2019dpod}. The most popular dataset for evaluation on the occlusion task has been Occlusion LineMOD \cite{brachmann2014learning}.

Depth information is crucial to recover the scale of the object since the scale is hard to recover from RGB images due to perspective projection. Therefore, 6D pose estimation from RGB images is an even more challenging problem. With the advent of deep learning approaches, 6D pose estimation purely from RGB images got more attention. The majority of the methods focus on initial detection and pose estimation. Some methods such as \cite{xiang2017posecnn} rely on end-to-end approach, while other leverage the object detection pipelines \cite{kehl2017ssd}. However, more accurate approaches rely on keypoint prediction \cite{hu2019segmentation, rad2017bb8} and compute pose via PnP algorithm \cite{lepetit2009epnp}. State-of-the-art methods \cite{zakharov2019dpod, peng2019pvnet} do not use hand-designed keypoints. They predict a correspondence map for every object pixel. Then the best minimal set of points is selected via the RANSAC algorithm. The 6D pose is computed via the PnP algorithm form the best set of points.  It is essential to notice that selecting a specific set of points is more beneficial than using all available points. Methods that focus on the occlusion task, such as \cite{hu2019segmentation}, explicitly select regions that are not occluded via segmentation mask. Excluding pixels that are not part of the object of interest always results in higher accuracy. 

In the domain of pose refinement, the most common method is ICP. It is an optimization-based method that iteratively improves the position by minimizing the distance between the model features and image data. Usually, the cost function minimizes the distance between the point cloud generated from the depth channel and 3D points of the object model. In the case of RGB data, the cost function minimizes the distance between the edges in the image and that of the model. Recently, few deep learning approaches such as DeepIM \cite{li2018deepim} and DPOD \cite{zakharov2019dpod} tackle the refinement task. Both approaches rely on pretrained models. DeepIM uses FlowNet \cite{fischer2015flownet} as the base model, which is designed for two image inputs. DPOD modifies the ResNet \cite{he2016deep} architecture to feed in two images. Furthermore, DeepIM boosts the performance by iteratively improving the prediction. The main idea of both papers is to reuse the weights trained on big datasets and transfer it to the 6D pose estimation task. However, these methods do not leverage the structure of the problem. Contrary, papers for initial pose estimation use advanced techniques, such as RANSAC for keypoint selection \cite{peng2019pvnet} or masking for removing occlusions \cite{hu2019segmentation}.

It has been shown that attention mechanisms improve the performance of deep neural network models in different domains such as machine translation \cite{gehring2017convolutional, vaswani2017attention}, visual recognition \cite{cao2019gcnet, wang2018non}, object segmentation \cite{huang2019ccnet}, tracking \cite{kosiorek2017hierarchical} and generative models \cite{zhang2019self}. The goal of attention mechanism is to extract the most relevant information form the feature vector. In the case of image data, a different attention map is applied for every query pixel separately \cite{wang2018non, huang2019ccnet, zhang2019self}, or the single attention map is used to extract global information \cite{cao2019gcnet, kosiorek2017hierarchical, jetley2018learn}. Although all these attention mechanisms bear similarity, the attention mechanism dramatically depends on the domain where it is applied. For example, using the attention model to establish a connection between each pixel and the rest of the image shows the best results on image segmentation task \cite{huang2019ccnet}, while using a single attention map performs better on visual recognition tasks \cite{cao2019gcnet}. In the 6D pose estimation area, the use of attention mechanisms has not been explored.

\section{Method}

\begin{figure*}
    \centering
    \includegraphics[width = 0.85 \textwidth]{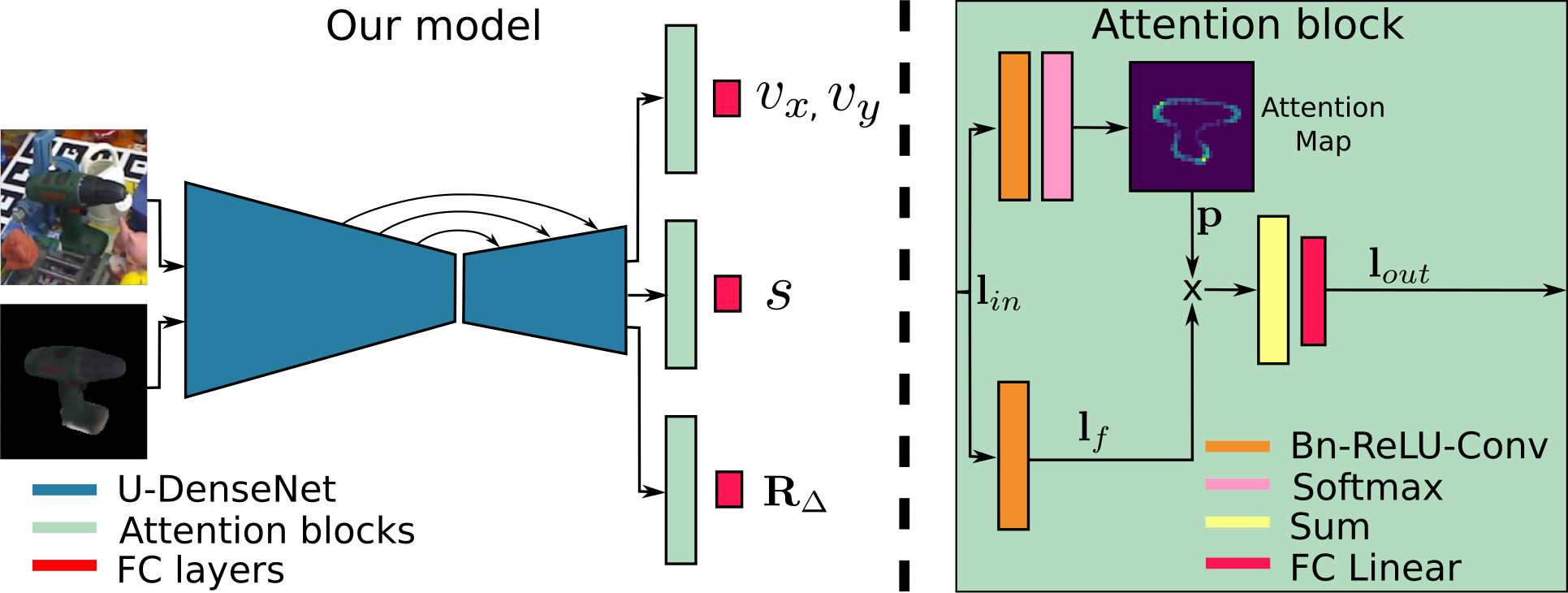}
    \caption{Architecture overview. Left: The network takes two input images and produces a set of relative pose update parameters ($x,y$ translation, scale $s$ and rotation $\mathbf{R}_{\Delta}$). Right: Details of the attention mechanism used in our model.}
    \label{fig:model}
\end{figure*}

We propose a novel neural network model for 6D pose refinement. Inspired by attention models in object detection and tracking \cite{jetley2018learn, kosiorek2017hierarchical}, we leverage an attention mechanism to improve the refinement of 6D pose estimation. 
We show that the spatial attention mechanism leads to the rejection of pixels from occluded object parts. Furthermore, the attention focuses only on a few sub-regions with discriminative details, which leads to better pose estimation overall. Importantly, we train the model for the downstream task of 6D pose estimation and do not use any specific loss terms to train the attention mechanism.

\subsection{Neural Network Model}

At the initial step, we obtain the object pose from an existing one-shot pose estimation algorithm, such as PVNet \cite{peng2019pvnet}. Our goal is to refine this initial object pose. The inputs to our network are two RGB images: the image crop around the object of interest, and the object rendered at the initial pose estimate (see Fig. \ref{fig:model}). To crop the image around the object, we project the object position to the image plane and select the crop area around the projection point. The rendered image provides initial pose information to the network. The main idea behind this setup is that the network learns to align these two images incrementally.

As the backbone of our model, we use DenseNet \cite{huang2017densely} (see Fig. \ref{fig:model}). More precisely, we leverage DenseNet variant designed for semantic segmentation \cite{jegou2017one}. The model uses skip connections to preserve detailed information in the up-stream. In our case, this is important since the attention model needs to detect fine spatial details. We use the full down-stream of the model from \cite{jegou2017one} and the first three blocks of the up-stream. Since the attention mechanism needs to identify image regions, we do not need the full resolution at the output. By using fewer layers we gain computational efficiency. Our backbone model already attains reasonable performance even when trained on the relatively small pose estimation datasets. This is due to the smaller number of parameters compared to ResNet \cite{he2016deep} or VGG \cite{simonyan2014very}. 

We separate the output into three streams: x-y displacement in the image plane $v_x, v_y$, scaling $s$, and rotation increments $\mathbf{R}_{\Delta}$ (see Fig. \ref{fig:model}). As these values relies on different kinds of information, it is advantageous to keep them separate. For example, x-y displacement requires only one point for perfect alignment, while scaling requires at least two points. For each stream we apply separate attention blocks, allowing each attention map to identify the most relevant features for their respective output. 

Our intention is to have an attention mechanism that can extract the discriminative details and reject the clutter, which is crucial for the 6D pose estimation task. 
Each attention block takes the output $\textbf{l}_{in} \in \mathbb{R}^{H \times W \times d_{in}}$ of the backbone and produces the output vector $\textbf{l}_{out} \in \mathbb{R}^{d_{out}}$. 
In the attention block, the backbone output $\textbf{l}_{in} \in \mathbb{R}^{H \times W \times d_{in}}$ is processed in two streams.  The first stream produces the attention tensor $\textbf{a} \in \mathbb{R}^{H \times W \times 1}$ whereas the second produces a feature tensor $\textbf{l}_{f} \in \mathbb{R}^{H \times W \times d_{f}}$. We apply a softmax activation function to the attention map $\textbf{a}$  which results in a 2D map of probabilities $\textbf{p} \in \mathbb{R}^{H \times W}$.
This probability map $\textbf{p}$ encodes the probability of a pixel in the input image being of high importance to the registration task. 
To extract these regions, we multiply $\textbf{p}$ element-wise with each slice $i$ of the tensor $\textbf{l}_{f}$ along the dimension $d_{f}$ and sum over the image dimensions:
\begin{equation}
    \textbf{l}^i = \sum_{W, H} \textbf{l}_{f}^i \cdot \textbf{p} \quad ,
\end{equation}
which produces a vector $\textbf{l} \in \mathbb{R}^{d_{f}}$. We feed the vector $\textbf{l}$ to the fully connected linear layer. The output of the linear layer gives the final attention output $\textbf{l}_{out}$ vector. The attention output is fed to a fully connected layer, which produces the network outputs (see Fig. \ref{fig:model}).

Since our starting point is an initial pose guess, the images of rendered and real objects typically already lie close to each other in the image plane. Thus, related details are relatively close but are not yet aligned. Our goal is to perfectly align all details. However, not all object parts contribute equal amounts of information. For example, homogeneous surfaces do not provide useful information for alignment. Leveraging spatial attention allows the network to identify those areas in the image that, with high probability, contribute meaningfully to the alignment. In other words, it identifies salient details, while rejecting clutter, occluded parts of the object and non-informative uniformly textured parts of the object. %

We evaluate our approach in a single-stage and multi-stage settings. The single-stage setting takes the output of the network described above as a final result. In the multi-stage setting, we take the output of the network and update the object pose in each stage. The new object pose is used as input to the next stage. Therefore, we stack four modules shown in Fig. \ref{fig:model} into a sequence, which enables us to train the network for iterative refinement. In both settings, our network performs better than state-of-the-art.

\subsection{Prediction Calculation}

To compute the prediction, we use the procedure from DeepIM \cite{li2018deepim}. The input to our model is the initial pose $[\mathbf{R_i}| \mathbf{t_i}]$, where  $\mathbf{R_i}$ is the initial object rotation matrix and $\mathbf{t_i}$ is the initial translation vector. We compute the pose increment $[\mathbf{R}_{\Delta}| \mathbf{t}_{\Delta}]$ from our network outputs. The final pose is obtained by applying pose increments to the initial pose $[\mathbf{R_f}| \mathbf{t_f}] = [\mathbf{R}_{\Delta}\mathbf{R_i}| \mathbf{t_i} +  \mathbf{t}_{\Delta}]$. 

The network directly outputs the rotation update $\mathbf{R}_{\Delta}$ (see Fig. \ref{fig:model}). In our case, the network predicts quaternions, which are then converted into a rotation matrix. To compute the translation increment $\mathbf{t}_{\Delta}$, we combine the two other outputs of the network. As suggested by DeepIM \cite{li2018deepim}, the network predicts the translation update in the image plane $(v_x, v_y)$ and image scaling factor $s$ separately. The translation update vector $\mathbf{t}_{\Delta}$ is the difference between the final translation vector $\mathbf{t_f} = [x_f, y_f, z_f]$ and the initial translation vector $\mathbf{t_i} = [x_i, y_i, z_i]$. The relation between these two vectors is given with the following equations:
\begin{equation}
\begin{split}
    v_x &= f_x(x_f/z_f - x_i/z_i), \\ 
    v_y &= f_y(y_f/z_f - y_i/z_i), \\ 
    s &= \log(z_i/z_f), \\ 
\end{split}
\end{equation}
where $f_x$ and $f_y$ are focal lengths defined by the intrinsic camera matrix. Using these relations, the final translation vector $\mathbf{t_f}$ can be computed.

\subsection{Training Loss}
Same as in other 6D pose refinement papers \cite{li2018deepim, zakharov2019dpod}, we use a training loss that computes the average euclidean distance between the object points transformed by ground truth transformation $[\mathbf{R}|\mathbf{t}]$ and predicted transformation $[\mathbf{R_f}|\mathbf{t_f}]$:
\begin{equation}\label{eq:loss_1}
	L_1 = \frac{1}{m} \sum_{\mathbf{x} \in \mathcal{M}} {|| (\mathbf{R}\mathbf{x} + \mathbf{t}) - (\mathbf{R_f}\mathbf{x} + \mathbf{t_f}) ||}_2 \quad ,
\end{equation}
where $\mathbf{x}$ are the model points in the set $\mathcal{M}$, which is a subset of CAD model vertices. The total number of points is $m$. In the refinement methods, the initial poses during training are generated by adding random noise to the ground truth pose. The separate training loss for symmetric objects is not used because the initial pose is closest to the ground truth pose among all symmetric poses. In the case of multi-stage prediction, we compute the loss after each stage. The final loss is the average loss over the four stages. %

\section{Implementation Details}

The original image is cropped around the center of the initial pose projection. As suggested in DeepIM, we crop the area $1.4$ times the size of the image object. It is then resized to 152x152 before being input into our refinement network. To generate the initial pose image, we use a differentiable render from \cite{henderson2019learning}. However, we do not use differentiable renderer functionality to backpropagate the gradients through the renderer. We tested this option, but there was no difference compared to the case when the gradients were blocked.

We train our network on a Pascal Titan X GPU and use the Stochastic Gradient Descent (SGD) optimizer. We train our single-stage model for $150$ epoch with learning rate $1 \times 10^{-2}$. The learning rate is then decreased to $1 \times 10^{-3}$ over next $50$ epochs. Further weight decay did not result in any improvement. We train the single-stage network from scratch with a batch size of $32$. To train the multi-stage network, we initialize each stage of the network with the weights from the single-stage model. We perform a warm-up to speed up the training time. In multi-stage setting, we use four stages of our model. We use batch size of $8$ due to GPU memory limitation.  We train our multi-stage model for $50$ epoch with learning rate $7 \times 10^{-3}$, when learning rate is decreased to $7 \times 10^{-4}$ and the model is trained for another $20$ epochs.

\subsection{Training Data}
We use the training data provided by the authors of PVNet, which consists of three datasets with objects from the LineMOD dataset. The first dataset contains real training data provided by the LineMOD dataset. The small number, roughly 1000 of images are sampled from the LineMOD sequences and are provided for training. The remainder of the dataset, around 13000 images, is used for testing only. However, this training dataset is too small to train deep neural networks. The second dataset contains 13000 synthetic images, 1000 for each object. To synthesize images, CAD models are rendered at different spatial positions, and the background and light sources are randomized. In this dataset, only a single object appears in the image without any occluders. The third dataset crops patches of LineMOD objects from real training data. These patches are then pasted over random backgrounds. When generating this dataset, multiple objects are placed in the same image, which can generate occlusions. This dataset contains 10000 images.

\begin{figure*}
    \centering
    \includegraphics[width = 0.95 \textwidth]{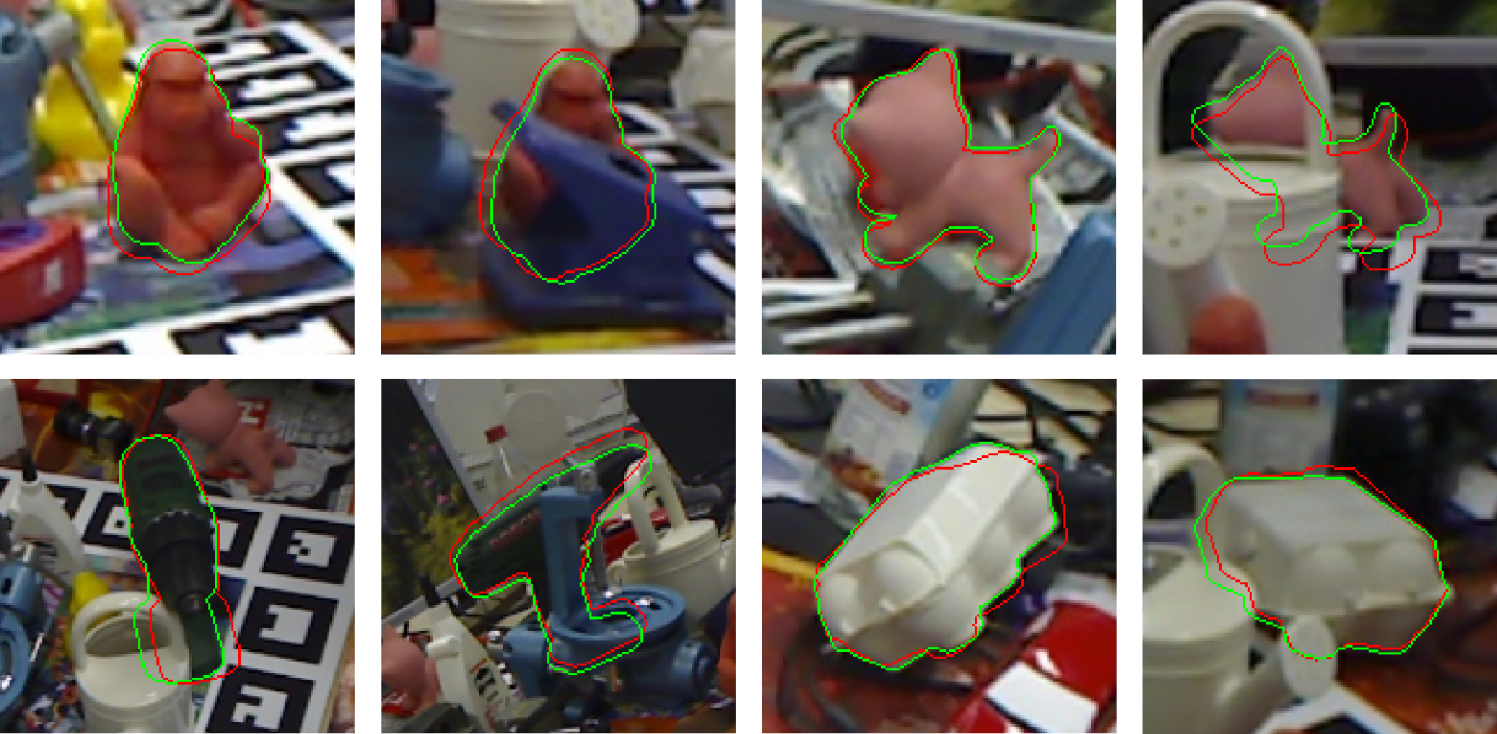}
    \caption{Examples of the object pose estimation. The red object outlines show the initial pose obtained from PVNet \cite{peng2019pvnet}. The green outlines indicate the pose obtained via our refiner.}
    \label{fig:examples}
\end{figure*}

The task of the refiner is to improve the pose that is already close to the ground truth pose. The examples of initialization poses are shown in Fig. \ref{fig:examples}. Thus, we can simulate the test scenario by adding random noise to ground truth poses. To generate the second input to the network at training time, we render the object at a randomized initial position. The rendered images are generated on the fly via a differentiable renderer. We generate randomized initial positions by adding noise to the ground truth position and feed them to the differentiable renderer.

\section{Results}
\label{sec:results}

In this section, we evaluate our approach in a series of experiments. We compare our approach with state-of-the-art methods and provide experimental evidence that the attention mechanism helps in the 6D pose refinement task. 

\subsection{Test Datasets}

Inline with previous work, in our experiments we use two datasets, namely LineMOD \cite{hinterstoisser2011gradient} and Occlusion LineMOD \cite{brachmann2014learning}. The LineMOD dataset is a standard dataset for the evaluation of 6D pose estimation methods. The dataset contains 13 sequences of 13 different objects recorded in a cluttered environment. In this dataset, objects are fully visible. To test 6D pose estimation in the presence of occlusions, the Occlusion LineMOD has been introduced \cite{brachmann2014learning}. This dataset contains 8 LineMOD objects that appear under occlusions.

\subsection{Evaluation Metric}

For evaluation, we use the standard ADD(-S) evaluation metric as is done in most 6D pose estimation papers \cite{peng2019pvnet, li2018deepim, zakharov2019dpod, kehl2017ssd, hinterstoisser2012model, brachmann2014learning}. ADD measures the average distance between the model points transformed by ground truth and the estimated poses. For symmetric objects, we use ADD-S, which is a modification of ADD that takes the symmetries of objects into account. In ADD-S the distance between the two closest points is used rather then measuring the euclidean distance between the all model points. In most of our results, we use a success rate threshold of $0.1d$ for ADD(-S), following related work.

\subsection{Baselines}
\label{sec:badelines}

We compare our model's performance with the SoA 6D pose refiners DeepIM \cite{li2018deepim} and DPOD \cite{zakharov2019dpod}. For DeepIM, we use the implementation provided by the authors. Since no publicly available implementation of DPOD exists, we use our own re-implementation, trained using the same data and randomization parameters as in our method. Moreover, we perform an ablation study to showcase the effect of the proposed attention model. To attain a baseline for the ablation, we train an additional model consisting of our backbone model, i.e., DenseNet \cite{huang2017densely}, and three output streams. In each output stream, we use two fully connected layers instead of the attention block. Since the ablation baseline does not predict attention maps, we use standard DenseNet instead of DenseNet for semantic segmentation \cite{jegou2017one}. Standard DenseNet has the optimal structure for regression tasks. The ablation baseline has the same number of parameters as our model. For more details about our baseline model, please refer to supplementary materials.

Clearly, the final pose estimate depends on the initial pose prediction. We use the SoA in the one-shot setting, PVNet \cite{peng2019pvnet}, for initialization. This allows for a comparison of refiners independently of the initialization method. Note that the settings reported in DPOD and DeepIM are incompatible (wrt to initialization), and hence we provide a new, directly comparable setting. The final performance of a refiner depends on the initialization (e.g., from PVNet). Thus, similar but not identical results, as reported in respective papers, are to be expected. On the LineMOD dataset, our baselines reach similar scores to ones reported in the original papers suggesting that our implementations are well tuned. On the Occlusion LineMOD dataset, the DPOD refiner performs slightly worse than reported in the paper. This is because in \cite{zakharov2019dpod}, a part of the test sequence is used as training data, which makes the task easier.  In the original Occlusion LineMOD settings \cite{brachmann2014learning}, the test sequence is not used for training. Both DeepIM \cite{li2018deepim} and our work uses the original setting.

\begin{table*}\centering
\scriptsize
\begin{center}
\begin{tabular}{|lc|ccc|c|ccc|c}\toprule
& & &Single stage & & & &Multi stage & \\\cmidrule{3-5}\cmidrule{7-9}
&PVNet init &DPOD &DenseNet & \textbf{Ours} & &DeepIM &DenseNet & \textbf{Ours} \\\midrule
ape &48.76 &70.67 &76.10 &\textbf{78.57} & &78.10 &76.95 &\textbf{80.76} \\
benchvise &99.03 &99.42 &\textbf{99.52} &\textbf{99.52} & &98.16 &99.32 &\textbf{99.61} \\
cam &87.06 &95.20 &\textbf{95.29} &93.73 & &95.49 &\textbf{97.16} &95.88 \\
can &96.85 &98.92 &98.92 &\textbf{99.41} & &98.03 &99.61 &\textbf{99.80} \\
cat &77.94 &90.92 &92.51 &\textbf{93.11} & &83.93 &93.41 &\textbf{94.31} \\
driller &96.53 &98.51 &98.32 &\textbf{98.71} & &95.44 &98.41 &\textbf{98.81} \\
duck &55.40 &76.90 &79.62 &\textbf{81.78} & &79.91 &82.25 &\textbf{85.26} \\
eggbox &99.72 &\textbf{100.00} &99.81 &99.81 & &90.80 &99.81 &\textbf{100.00} \\
glue &82.24 &83.59 &83.59 &\textbf{83.69} & &78.47 &83.49 &\textbf{83.78} \\
holepuncher &79.64 &87.73 &\textbf{91.53} &90.58 & &46.81 &89.34 &\textbf{90.10} \\
iron &98.37 &98.98 &98.77 &\textbf{99.08} & &99.18 &\textbf{99.59} &98.77 \\
lamp &99.33 &99.90 &\textbf{100.00} &\textbf{100.00} & &98.46 &99.90 &\textbf{100.00} \\
phone &91.45 &\textbf{97.60} &97.31 &97.12 & &89.91 &97.98 &\textbf{98.56} \\
\midrule
MEAN &85.56 &92.18 &93.18 &\textbf{93.47} & &87.13 &93.63 &\textbf{94.28} \\
\bottomrule
\end{tabular}
\end{center}
\caption{Pose estimation performance on the LineMOD dataset. The table shows percentage of correctly estimated poses using $0.1d$ ADD(-S) threshold (higher is better). For initialization, we use 6D poses obtained from PVNet (the performance of the initial poses is shown in the left-most column). Since the initialization points are different compared to the original papers, similar but not identical results are to be expected. Our method performs comparable or better than prior work and our DenseNet baseline for most objects and outperforms the SoA on average.}\label{tab:add_linemod}
\end{table*}

\subsection{LineMOD Results}

\begin{table}\centering
\scriptsize
\begin{center}
\resizebox{0.95\linewidth}{!}{
\begin{tabular}{lcccccccc}\toprule
& &DenseNet & & & & \textbf{Ours} & \\\cmidrule{2-4}\cmidrule{6-8}
&0.1 d &0.05 d &0.02 d & &0.1 d &0.05 d &0.02 d \\\midrule
ape &76.95 &48.00 &\textbf{11.62} & &\textbf{80.76} &\textbf{50.76} &8.57 \\
benchvise &99.32 &88.28 &41.18 & &\textbf{99.61} &\textbf{91.18} &\textbf{46.12} \\
cam &\textbf{97.16} &\textbf{77.94} &\textbf{25.29} & &95.88 &76.57 &24.51 \\
can &99.61 &88.88 &39.67 & &\textbf{99.80} &\textbf{89.27} &\textbf{40.55} \\
cat &93.41 &70.86 &25.65 & &\textbf{94.31} &\textbf{71.36} &\textbf{26.75} \\
driller &98.41 &87.91 &43.31 & &\textbf{98.81} &\textbf{92.67} &\textbf{49.45} \\
duck &82.25 &51.64 &\textbf{12.39} & &\textbf{85.26} &\textbf{55.40} &12.11 \\
eggbox &99.81 &90.23 &30.05 & &\textbf{100.00} &\textbf{93.15} &\textbf{40.19} \\
glue &83.49 &75.87 &34.85 & &\textbf{83.78} &\textbf{76.64} &\textbf{38.22} \\
holepuncher &89.34 &54.23 &5.04 & &\textbf{90.10} &\textbf{58.04} &\textbf{13.23} \\
iron &\textbf{99.59} &88.25 &42.08 & &98.77 &\textbf{89.17} &\textbf{44.33} \\
lamp &99.90 &90.31 &30.04 & &\textbf{100.00} &\textbf{95.01} &\textbf{48.18} \\
phone &97.98 &78.19 &30.26 & &\textbf{98.56} &\textbf{80.50} &\textbf{34.39} \\
\midrule
MEAN &93.63 &76.20 &28.57 & &\textbf{94.28} &\textbf{78.44} &\textbf{32.82} \\
\bottomrule
\end{tabular}
}
\end{center}
\caption{ADD(-S) scores for various thresholds on the LineMOD dataset (higher is better). We compare the multi-stage variant of our approach to the best baseline by tightening the success rate threshold. Our method predicts poses that are closer to the ground truth pose and the gap to the baseline increases for lower thresholds.}\label{tab:ADD_all}
\end{table}

We first evaluate our approach on the LineMOD dataset \cite{hinterstoisser2011gradient}. The results are shown in Table \ref{tab:add_linemod}. For each of the objects in the LineMOD dataset, we compute the ADD(-S) scores. The DPOD method is a single-stage refiner, while the DeepIM uses multiple stages. Thus, we train our model and the DenseNet baseline in both configurations. We show qualitative results of our method in Fig.~\ref{fig:examples}.

As presented in Table \ref{tab:add_linemod}, our approach shows the best performance in both of the settings. Furthermore, the vanilla DenseNet performs better than both SoA refiners. This confirms our hypothesis that the DenseNet backbone is an effective substitute for the ResNet based architectures used in prior work, despite a much smaller footprint in terms of parameters. Finally, adding our proposed attention mechanism shows a consistent increase in performance over DenseNet, which confirms the effectiveness of the attention mechanism.

To further evaluate the influence of our contribution, we provide the ADD(-S) score for thresholds $0.05d$ and $0.02d$, similar to \cite{li2018deepim}, which makes the task increasingly harder. The results are presented in Table \ref{tab:ADD_all}. As the threshold goes down, the difference between our approach and the DenseNet baseline becomes more evident. The attention blocks helps to isolate the most salient features for matching, which allows the refinement network to achieve higher accuracy.

To verify that these results are not only a side-effect of our experimental setting we also provide results using the initial pose estimates used in DeepIM. Note that here the accuracy of the initialization method is significantly lower than PVNet \cite{peng2019pvnet} which implies that some initial estimates maybe outside of the distribution used during training of our method. The results in Table \ref{tab:ADD_deepIM} again show that our method significantly improves over DeepIM, indicating that the performance increase is a direct result of the proposed attention mechanism.

\subsection{Occlusion LineMOD Results}

To evaluate our approach in settings that contain occlusions, we perform a similar experiment on the Occlusion LineMOD dataset. The results are presented in Table \ref{tab:ADD_occlusion}. We use PVNet to obtain initial positions and evaluate our approach using ADD(-S) metric. We show the performance of DPOD and DeepIM refiners as well using the same initial positions. For qualitative results refer to Fig.~\ref{fig:examples}.

Table \ref{tab:ADD_occlusion} shows that our approach performs better than the baselines. The single-stage variant achieves a significant improvement over DenseNet. This is due to the attention map being shaped such that pixels of the occluder are ignored, which has been shown to improve accuracy \cite{hu2019segmentation}. The difference is more pronounced in the multi-stage case.

\begin{table}\centering
\scriptsize
\begin{center}
\begin{tabular}{lrrrrr}\toprule
&Init &DeepIM &DenseNet &\textbf{Ours} \\\midrule
MEAN &62.7 &88.6 &85.17 &\textbf{92.02} \\
\bottomrule
\end{tabular}
\end{center}
\caption{ADD(-S) scores for initialization points used in DeepIM on the LineMOD dataset (higher is better). We compare all multi-stage methods. The DeepIM performance is the same as reported in the DeepIM paper since we use the original initialization points.  Ours outperforms the SoA on average.}\label{tab:ADD_deepIM}
\end{table}

\begin{table*}\centering
\scriptsize
\begin{center}
\begin{tabular}{|lc|ccc|c|ccc|c}\toprule
& & &Single stage & & & &Multi stage & \\\cmidrule{3-5}\cmidrule{7-9}
&PVNet init &DPOD &DenseNet &\textbf{Ours} & &DeepIM &DenseNet &\textbf{Ours} \\\midrule
ape &15.98 &19.91 &28.29 &\textbf{28.72} & &\textbf{41.71} &33.68 &38.29 \\
can &63.79 &51.86 &52.44 &\textbf{55.34} & &62.30 &68.43 &\textbf{75.39} \\
cat &19.38 &16.76 &15.75 &\textbf{21.65} & &24.94 &23.42 &\textbf{25.27} \\
driller &64.25 &\textbf{71.58} &64.42 &66.14 & &59.06 &72.16 &\textbf{77.18} \\
duck &31.81 &40.32 &40.23 &\textbf{44.17} & &41.64 &44.26 &\textbf{47.33} \\
eggbox &41.02 &44.34 &43.91 &\textbf{44.77} & &\textbf{50.21} &48.34 &49.36 \\
glue &41.42 &\textbf{47.84} &43.63 &\textbf{47.84} & &\textbf{55.59} &48.06 &51.38 \\
holepuncher &39.08 &37.15 &43.85 &\textbf{45.61} & &43.14 &45.77 &\textbf{48.79} \\
\midrule
MEAN &39.59 &41.22 &41.57 &\textbf{44.28} & &47.33 &48.02 &\textbf{51.63} \\
\bottomrule
\end{tabular}
\end{center}
\caption{Pose estimation performance on the Occlusion LineMOD dataset. The table shows percent of correctly estimated poses using $0.1d$ ADD(-S) threshold (higher is better). As initialization points, we use 6D poses obtained from PVNet (the performance of the initial poses is shown in the left-most column). Since the initialization points are different compared to the original papers, similar but not identical results are to be expected. The performance of DPOD method is slightly worse compared to the results reported in the paper due to the different training settings used in the DPOD paper (refer to Sec.~\ref{sec:badelines} for more details). Ours significantly outperforms the previous SoA.}\label{tab:ADD_occlusion}
\end{table*}

\subsection{Analysis of the Attention Model}
\label{sec:att_model}

\begin{figure*}
    \centering
    \includegraphics[width = 0.95 \textwidth]{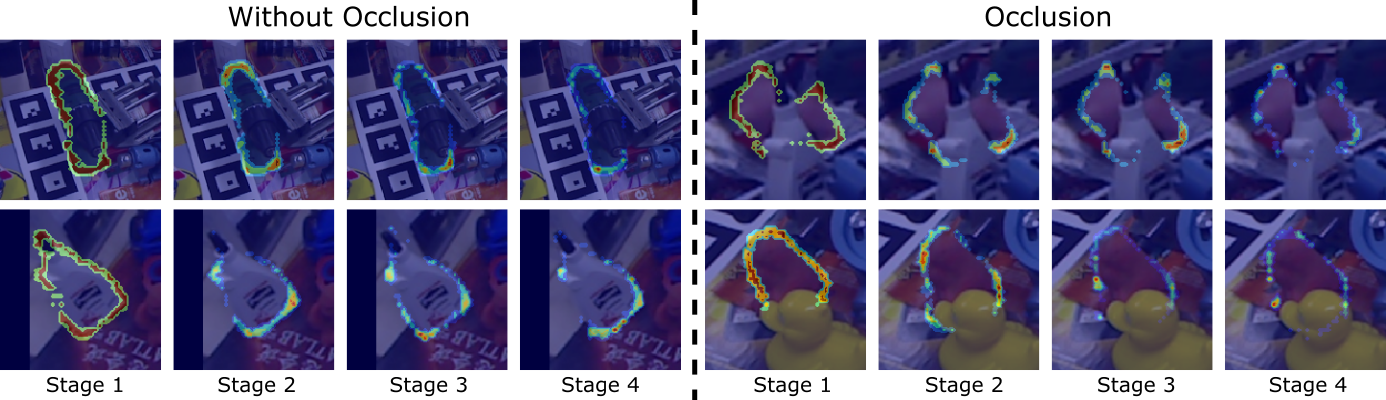}
    \caption{We overlay the attention probability map over the input image to show which details the attention focuses on. We plot the probability map at each stage of our model. The attention isolates increasingly specific features in the later stages. For example, the object parts that overlap with the white background are avoided on the glue example (bottom left). Furthermore, the attention avoids occlusions (e.g., the rubber duck, bottom right).}
    \label{fig:att}
\end{figure*}

To better understand the effectiveness of our approach, we analyze the outputs of the attention model. We generate a heatmap based on the predicted attention map and overlay it over the input image, some examples are shown in Fig.~\ref{fig:att}.

First, Fig.~\ref{fig:att} illustrates that the attention mostly focuses on the outline of the object. The refiner does not need to consider global features because the initial object pose is close to the ground truth pose. Global features need to be considered only in cases where initial displacement is significant. For example, if the initial rotation is around 180 degrees, the network needs to distinguish between the front and back of the object. Since the initial displacement is small, the network focuses on overlaying the characteristic features, which are usually located on the outline of the object. This is a somewhat expected result, since object outline edges are also a common choice for ICP-based refinement from RGB images. 

Second, we observe that the attention maps tend to get even more focused on specific regions in the later stages of refinement. As the pose estimate converges to the real image, only a few spatial features need to be considered  to improve the prediction. Finally, we notice that the attention map avoids occluded object parts (see Fig. \ref{fig:att}, right). Please keep in mind that the attention map is learned without any direct supervision and is shaped only via directly optimizing the pose estimation task.

\section{Conclusion}

In this paper, we present a novel attention-based model method for 6D pose refinement. Our model uses an attention mechanism to recognize the regions containing details that give more accurate pose estimation. For the refinement task, details are more important than high-level features since the rendered model is already close to the goal pose. Thus, the network can focus on small pose increments by matching details of the model and input image. We experimentally show that the attention focuses on the outline regions, which contains plenty of discriminative details. In the ablation study, we demonstrate that the attention mechanism leads to performance improvement on different evaluation datasets and using different initialization poses. We experimentally show that our method leads to a significant improvement over SoA results.

To the best of our knowledge, this is the first time the attention mechanism is used for the 6D pose estimation task. Based on our experiments, we believe that the attention mechanism has particular importance for the task of 6D pose estimation. We hope that our insights will facilitate research in this direction. For example, one possibility is to use the attention mechanism for feature selection instead of the RANSAC algorithm. Furthermore, one can use attention to eliminate the influence of the occlusion object even in cases when the segmentation labels are not provided.

\paragraph*{Acknowledgments}
We thank the NVIDIA Corporation for the donation of GPUs used in this work.

{\small
\bibliographystyle{ieee}
\bibliography{egbib}
}

\clearpage
\section{Appendix}

\subsection{Model Details}

Here we describe the architecture of our model and the DenseNet baseline in more details. The core of the DenseNet architecture is a Dense block \cite{huang2017densely}. The block uses convolutional layers of size 3x3. The input to the convolutional layer is first processed via batch normalization and ReLU activation. The main characteristic of the Dense block is that new features obtained from the latest convolutional layer are concatenated to the layer input and propagated further. As a result, DenseNet keeps the original features and only adds new features to the feature stack. Besides the Dense block, DenseNet has transition layers to reduce the resolution of the input features. We use a 2x2 max pooling layer with stride 2 as a transition layer.

\begin{table}[!htp]\centering

\vspace{0.5cm}
\begin{tabular}{|c|}
\hline
Ours \\
\hline\hline
Input, $m = 3$ \\
\hline
3 x 3 Convolution, $m = 48$ \\
\hline
DB (4 layers) + TD, $m = 112$ \\
\hline
DB (5 layers) + TD, $m = 192$ \\
\hline
DB (7 layers) + TD, $m = 304$ \\
\hline
DB (10 layers) + TD, $m = 464$ \\
\hline
DB (12 layers) + TD, $m = 656$ \\
\hline
DB-up (15 layers), $m = 896$ \\
\hline
TU + DB-up (12 layers), $m = 1088$ \\
\hline
TU + DB-up (10 layers), $m = 816$ \\
\hline
TU + DB (7 layers), $m = 578$ \\
\hline
Attention block (3 streams) \\
\hline
FC ($d$ = Out dimension) (3 streams) \\
\hline
\end{tabular}
\vspace{0.5cm}
\caption{
The architecture details of our model. DB is Dense Block, TD is Transition Down block, TU is Transition Up block, DB-up is modified Dense Block used in up-stream, and FC is a fully-connected layer. Each layer in DB and DB-up increases the number of features by 16. The number $m$ is the total number of features used in the DB and DP-up blocks.  $d$ is the dimension of the fully connected layers. }\label{tab:ours_arch}
\end{table}

DenseNet for semantic segmentation \cite{jegou2017one} uses the DenseNet architecture in the down-stream, but modifies the architecture slightly in the up-stream. First, the features are upsampled in the up-stream via the transition layer that uses 3x3 transposed convolutions with stride 2. The network uses the skip connections to add features with the same spatial resolution from the down-stream to the upsampled features from the previous layer. These features are inputs to the up-stream Dense block. Since the spatial resolution increases, the network will require too much memory if we continue to concatenate the feature vectors. Thus, the output of the Dense block in the up-stream is modified (DB-up in Table \ref{tab:ours_arch}). The block output consists only of the features created in the current Dense block, i.e., the new features are not concatenated to the input features. In the up-stream, we can access upsampled features, down-stream features with the same resolution, and newly generated features. We use all these features as the input to the attention block. We show details of our model in Table \ref{tab:ours_arch}.

\begin{table}[!htp]\centering
\begin{tabular}{|c|}
\hline
DenseNet \\
\hline\hline
Input, $m = 3$ \\
\hline
3 x 3 Convolution, $m = 48$ \\
\hline
DB (4 layers) + TD, $m = 112$ \\
\hline
DB (5 layers) + TD, $m = 192$ \\
\hline
DB (14 layers) + TD, $m = 416$ \\
\hline
DB (20 layers) + TD, $m = 736$ \\
\hline
DB (21 layers) + TD, $m = 1056$ \\
\hline
DB (15 layers), $m = 1296$ \\
\hline
Global average pooling \\
\hline
FC ($d = 512$) + Batch Norm + ReLU (3 streams) \\
\hline
FC ($d =$ Out dimension) (3 streams) \\
\hline
\end{tabular}
\vspace{0.5cm}
\caption{The architecture details of our ablation baseline DenseNet. DB is Dense Block, TD is Transition Down block, and FC is a fully-connected layer. Each layer in DB increases the number of features by 16. The number $m$ is the total number of features used in the DB blocks. $d$ is the dimension of the fully connected layers.}\label{tab:dn_arch}
\end{table}

\begin{table*}\centering
\scriptsize
\begin{center}
\begin{tabular}{|l|ccc|c|ccc|}\toprule
&  &Single stage & & & &Multi stage & \\\cmidrule{2-4}\cmidrule{6-8}
&DPOD &DenseNet &\textbf{Ours} & &DeepIM &DenseNet &\textbf{Ours} \\\midrule
\midrule
fps &80 &50 &35 & &9 &17 &13 \\
\bottomrule
\end{tabular}
\end{center}
\caption{Results of the runtime experiment. We compare the running times of our methods in the single-stage and multi-stage settings. All experiments are done on a Pascal Titan X GPU.}\label{tab:fps}
\end{table*}

\begin{figure*}
    \centering
    \includegraphics[width = 0.9 \textwidth]{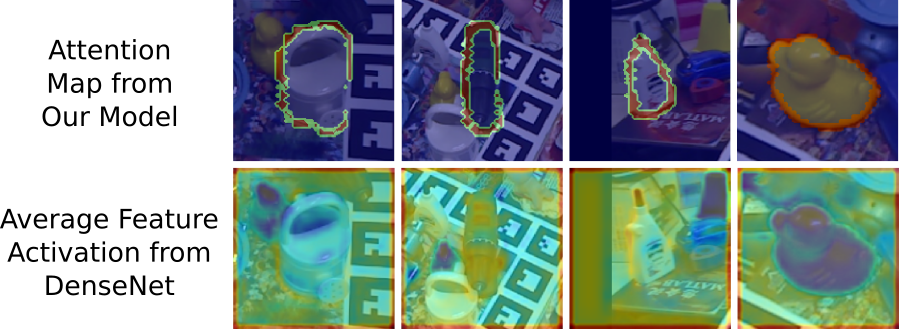}
    \caption{In this figure, we compare our model's attention map to the average feature activation from the third DenseNet block (the block with 14 layers in Table \ref{tab:dn_arch}). We use results from the first stages. The feature activations look very similar in later stages. Red and yellow colors indicate regions with higher activation and attention values, while blue represents regions with low values. Feature activations have high values at outlines, but also in other regions of the input image. The attention helps to isolate the regions that are relevant to the target object and reject the clutter.}
    \label{fig:att_vs_activation}
\end{figure*}

Our ablation baseline is shown in Table \ref{tab:dn_arch}. We use regular DenseNet since the baseline directly predicts the rotation and translation parameters, and hence does not need to upscale feature vectors. 
To have the same number of parameters in both models, we increase the number of layers Dense blocks 3, 4, and 5 (the Dense blocks with 14, 20, and 21 layers in Table \ref{tab:dn_arch}). We choose these blocks since our model has more capacity at these resolutions. To obtain the strong ablation baseline, we experimented with the number and size of fully connected layers at the end of the networks. There was no improvement when we added more fully connected units or additional layers.

\subsection{Runtime Experiment}

We compared the runtime of our method to all baseline models. We present the results in Table \ref{tab:fps}. Since all models are fully differentiable, the computation time depends on the neural network size. In single-stage settings, all methods run faster than 30 fps. Most of the modern cameras operate at 30 fps, and hence we can say that all methods operate in real-time. In the multi-stage setting, all methods operate at a similar frame rate. However, we show in the paper that our method outperforms all baseline methods. In conclusion, our method performs better without sacrificing computational time.

\subsection{Attention vs. Activation Experiment}

In Fig. \ref{fig:att_vs_activation}, we compare the attention maps of our model to the average feature activation of the DenseNet baseline. The resolution of our attention maps is the same as the resolution of the features in the Dense block with 14 layers in Table \ref{tab:dn_arch}. Thus, we use the features at the output of this block. We compute per-pixel average values. DenseNet activation maps look very similar in all four stages of the multi-stage model. We show results from the first stages, but we can reach the same conclusions by comparing any of the stages. 

Similarly to the attention model, features activations have high values at object outlines. However, these are not the only regions where activations have high values. The activations are high at the background regions and even at the uniformly colored parts of the object (see driller example in Fig. \ref{fig:att_vs_activation}). One can argue that deeper layers of DenseNet can potentially reject some of the clutter. However, deeper layers have lower resolutions, which prevents them from isolating fine-grained details. The attention network resolves this issue by enabling our model to isolate discriminative details at the high resolution. This leads to accuracy improvement, as we show in our paper.

\subsection{U-DenseNet Ablation Baseline Experiment}

In this experiment, we compare our results to an additional ablation baseline. In the Sec.~\ref{sec:results}, we use DenseNet as our ablation baseline because the downstream type of architecture is the most typical choice for a regression task. The features need to be squeezed into a one-dimensional vector to perform regression in the final part of the network. If we do not need access to low-level features, there is no reason to use the up-stream part of the network. In our model, we apply a spatial attention mechanism for feature selection, and thus we need the up-stream architecture. To confirm that the specific backbone details are not the main cause for performance improvement, we conduct an additional experiment. We trained a new ablation model, where we use the same U-DenseNet backbone as in our model. To obtain the new ablation baseline, we replace the attention block with fully connected layers. We use global average pooling to compute one-dimensional feature vectors for regression. 

\begin{table*}[!htp]\centering
\scriptsize
\begin{center}
\begin{tabular}{lcccccccc}\toprule
& &LineMOD & & & &Occlusion LineMOD& \\\cmidrule{2-4}\cmidrule{6-8}
&U-DenseNet &DenseNet &Ours & &U-DenseNet &DenseNet &Ours \\\midrule
Single-stage &92.77 &93.18 &\textbf{93.47} & &42.04 &41.57 &\textbf{44.28} \\
Multi-stage &93.70 &93.63 &\textbf{94.28} & &49.27 &48.02 &\textbf{51.63} \\
\bottomrule
\end{tabular}
\end{center}
\caption{Average ADD(-S) scores of our approach compared to DenseNet and U-DenseNet ablation baselines on LineMOD and Occlusion LineMOD datasets. The table shows the percentage of correctly estimated poses using $0.1d$ ADD(-S) threshold (higher is better). Our method performs better than both baselines. }\label{tab:add_udn}
\end{table*}

\begin{table*}[!htp]\centering
\scriptsize
\begin{center}
\begin{tabular}{lcccccc}\toprule
&PVNet init &One Stage &Two Stages &Three Stages &Four Stages \\\midrule
LineMOD &85.56 &93.47 & 94.15 & 94.19 & \textbf{94.28} \\
Occlusion LineMOD &39.59 &44.28 &49.89 &51.21 &\textbf{51.63} \\
\bottomrule
\end{tabular}
\end{center}
\caption{Average ADD(-S) scores of our approach for a different number of stages. The table shows the percentage of correctly estimated poses using $0.1d$ ADD(-S) threshold (higher is better). We present average scores on the LineMOD and Occlusion LineMOD datasets.}\label{tab:add_stages}
\end{table*}

We present results in Table \ref{tab:add_udn}, where U-DenseNet represents the new baseline. By comparing the accuracies of the ablation baselines, i.e., the U-DenseNet and DenseNet baselines, we conclude that both models perform similarly. U-DenseNet performs slightly better compared to the DenseNet on the Occlusion LineMOD dataset. U-DenseNet has direct access to the not occluded low-level features, which is probably the cause of this difference. However, our approach performs significantly better compared to both ablation baselines, which confirms that the performance improvement is due to the attention module.

\subsection{Influence of Stages Number}

The multi-stage variant of our approach uses four stages. We follow the DeepIM paper in this regard. For a fair comparison, we use the same number of stages as the multi-stage baseline. To evaluate how our method performs with a different number of stages, we conduct additional experiments. In this experiment, we train our model using two and three stages and compare to the models evaluated in the main paper, i.e., the single-stage and four-stage models.  We present the results in Table~\ref{tab:add_stages}.

Table~\ref{tab:add_stages} shows that adding each stage benefits the refinement procedure. The models with one and two stages give the most significant improvements. The later stages still improve the results, but the relative improvement is smaller compared to the first two stages. From the improvement trend in Table \ref{tab:add_stages}, we can conclude that adding more than four stages would improve the results but not significantly. It would be interesting to increase the number of stages to find the saturation point. However, this is not possible due to hardware limitations. We leave this for future work.

\subsection{Attention Model Experiment}

\begin{figure*}
    \centering
    \includegraphics[width = 0.8 \textwidth]{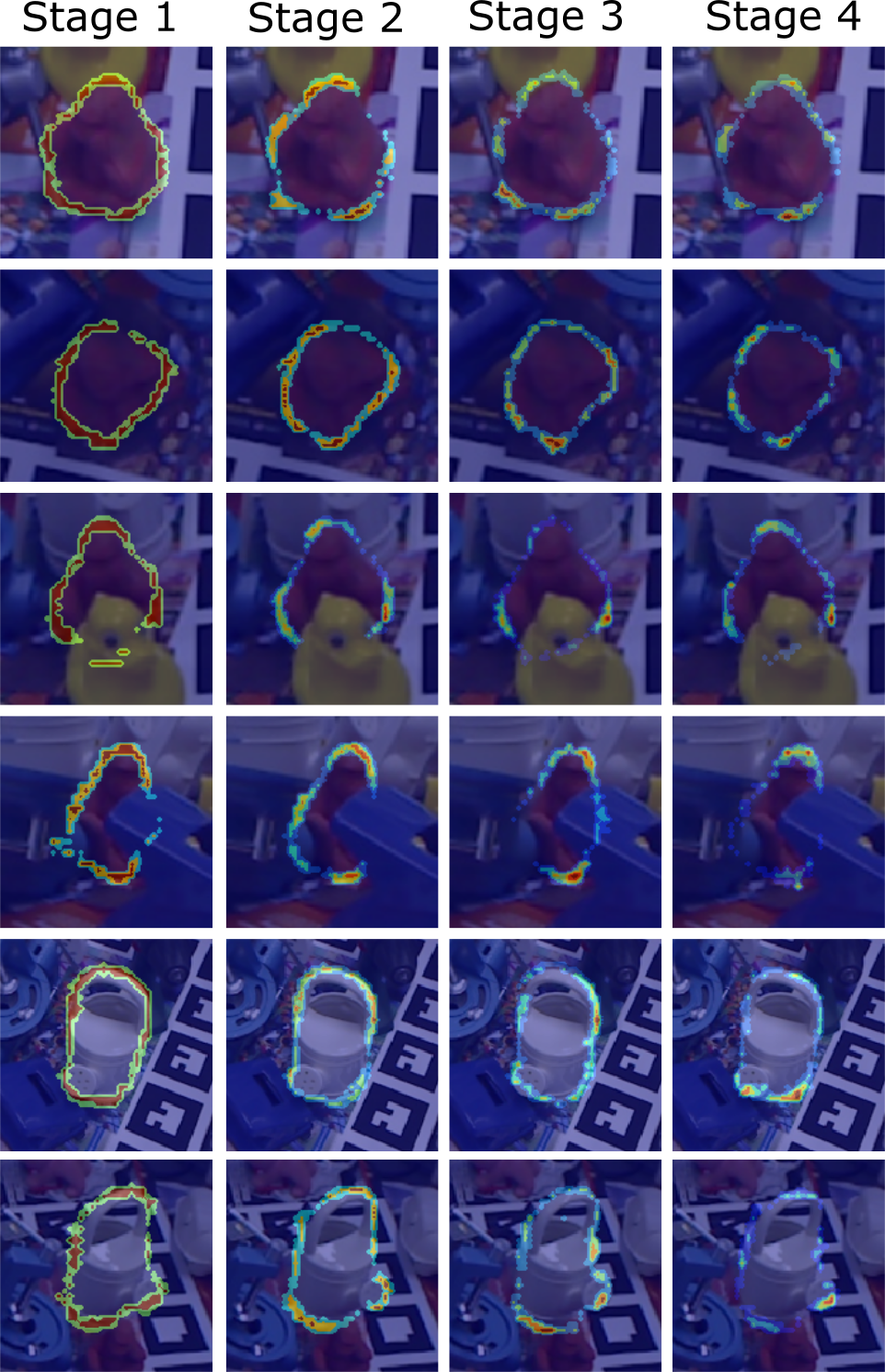}
    \caption{We overlay the attention map on the input image to indicate the regions that the attention model extracts. We plot the attention map after each stage of our model. The attention focuses on discriminative details and avoids occlusions. In later stages of the network, the attention isolates increasingly specific features. }
    \label{fig:fig_1}
\end{figure*}

\begin{figure*}
    \centering
    \includegraphics[width = 0.8 \textwidth]{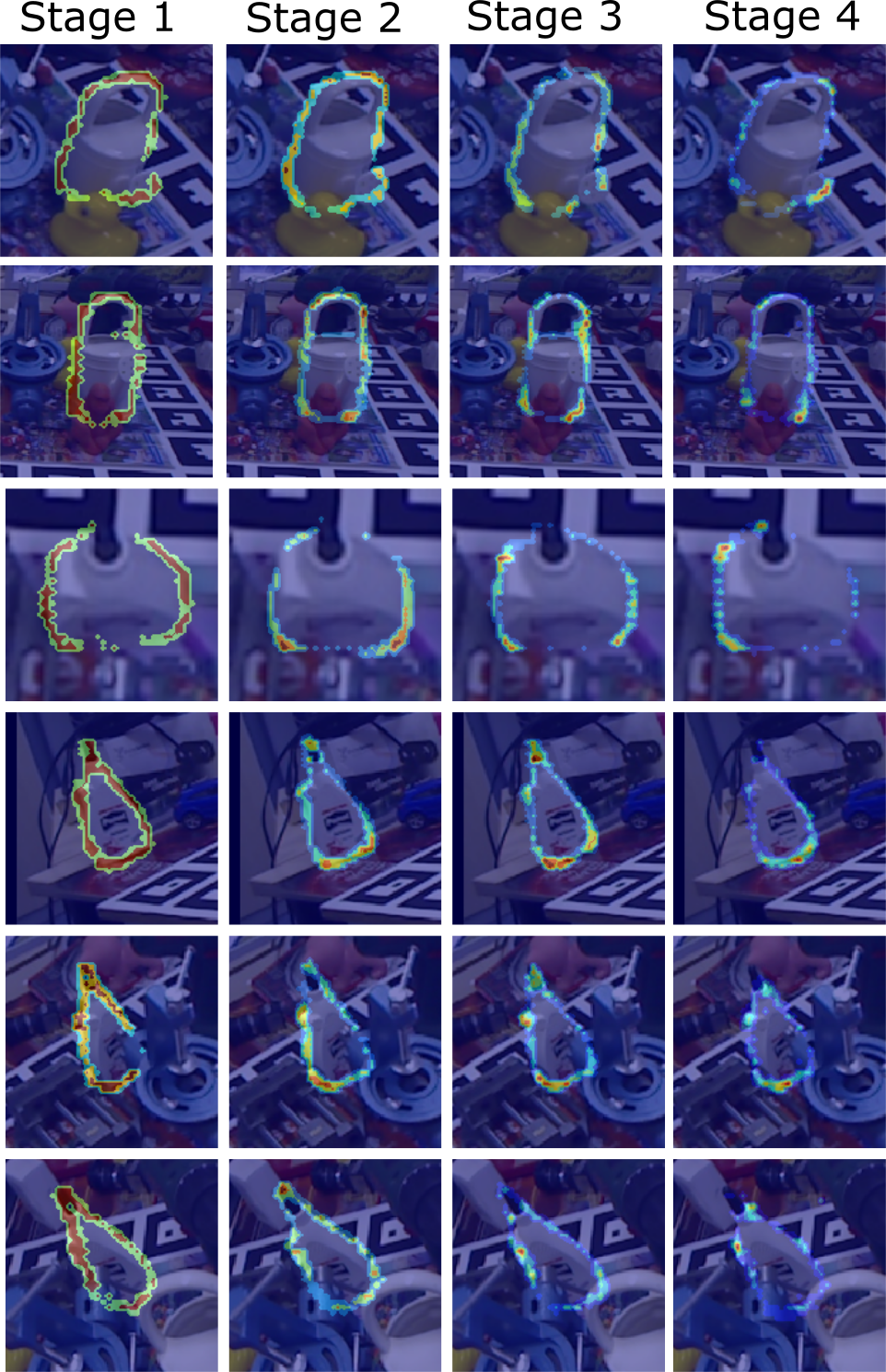}
    \caption{We overlay the attention map on the input image to indicate the regions that the attention model extracts. We plot the attention map after each stage of our model. The attention focuses on discriminative details and avoids occlusions. In later stages of the network, the attention isolates increasingly specific features. }
    \label{fig:fig_2}
\end{figure*}

\begin{figure*}
    \centering
    \includegraphics[width = 0.8 \textwidth]{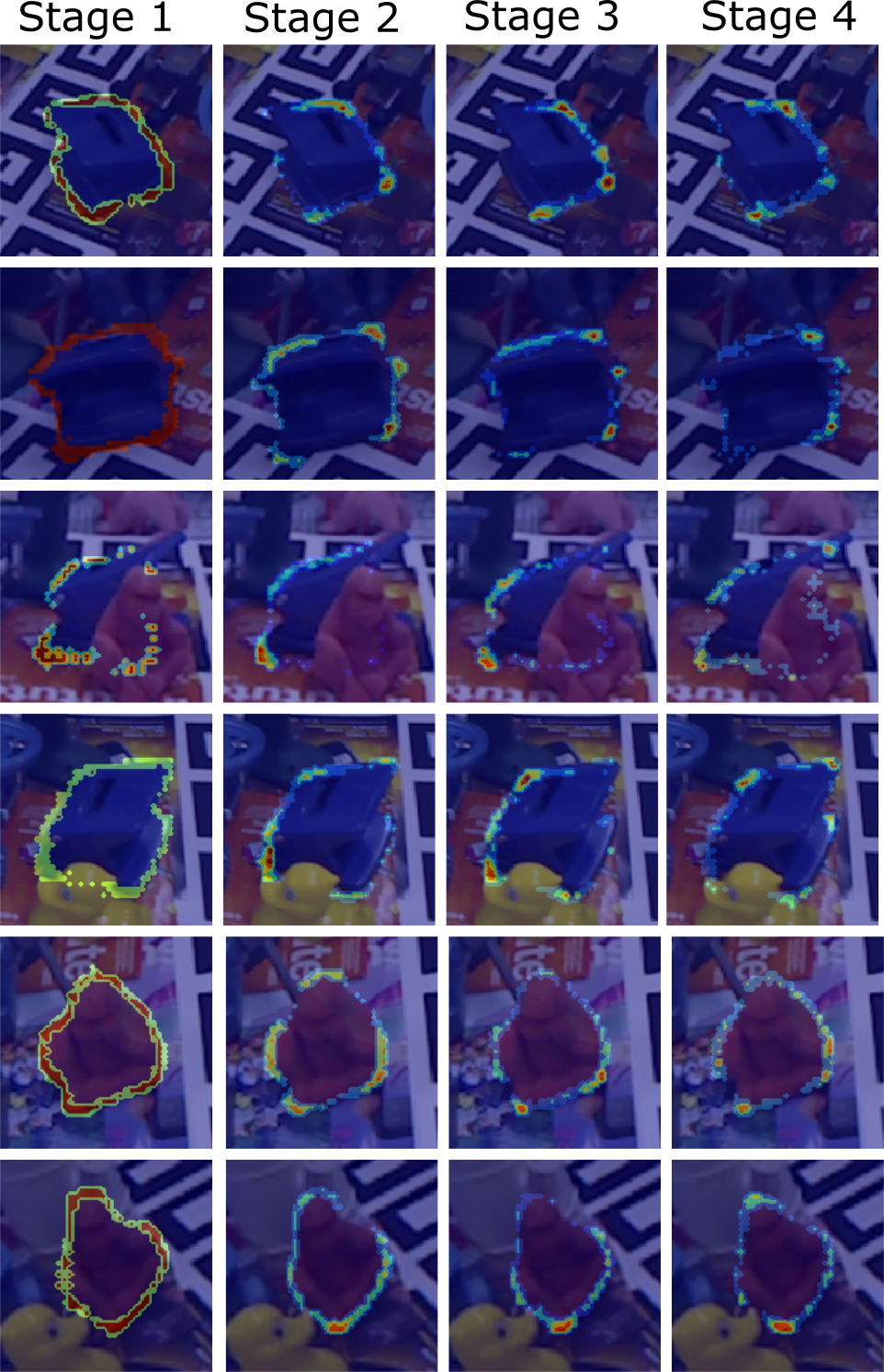}
    \caption{We overlay the attention map on the input image to indicate the regions that the attention model extracts. We plot the attention map after each stage of our model. The attention focuses on discriminative details and avoids occlusions. In later stages of the network, the attention isolates increasingly specific features. }
    \label{fig:fig_3}
\end{figure*}

In this section, we provide additional results to evaluate our attention model. Same as in Sec. 5.6 in our paper, we plot attention maps over the real-world image. We show attention maps after each stage in Fig.~\ref{fig:fig_1}, Fig.~\ref{fig:fig_2}, Fig.~\ref{fig:fig_3}. Here we confirm that the attention focuses on the object outline while avoiding the occlusions. Furthermore, the attention isolates more specific details in the later stages. We confirm our conclusion from Sec. 5.6 in the paper on the larger set of test images.

\subsection{Iterative Refinement Qualitative Experiment}

\begin{figure*}
    \centering
    \includegraphics[width = 0.9 \textwidth]{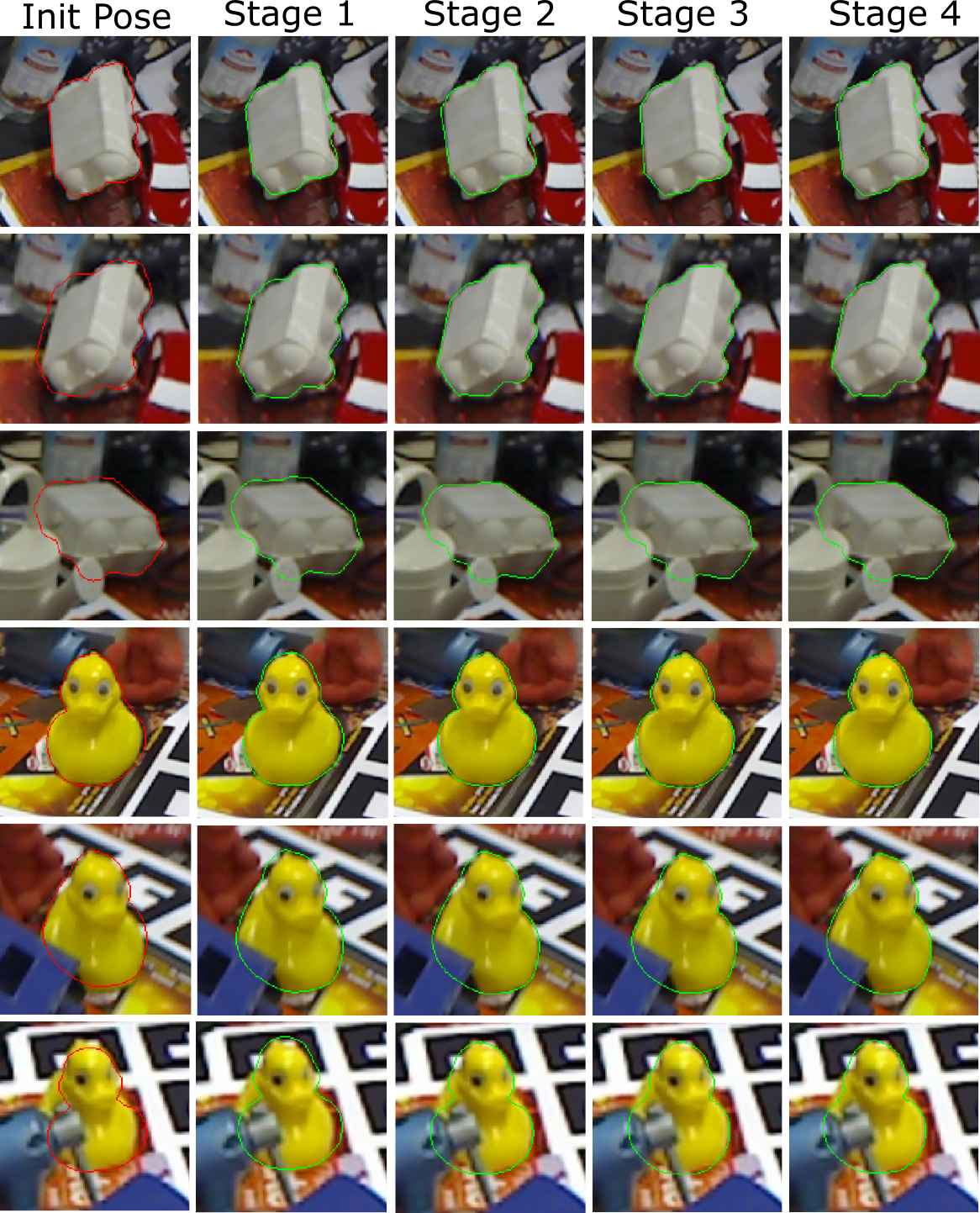}
    \caption{Iterative refinement results. For initialization, we use PVNet \cite{peng2019pvnet}.  The red object outlines show the initial poses. The green outlines indicate the pose obtained via our refiner. We show the pose obtained after each step of the refiner. Notice that after each iteration, the rendered outline is closer to the real object outline.
}
    \label{fig:fig_4}
\end{figure*}

\begin{figure*}
    \centering
    \includegraphics[width = 0.9 \textwidth]{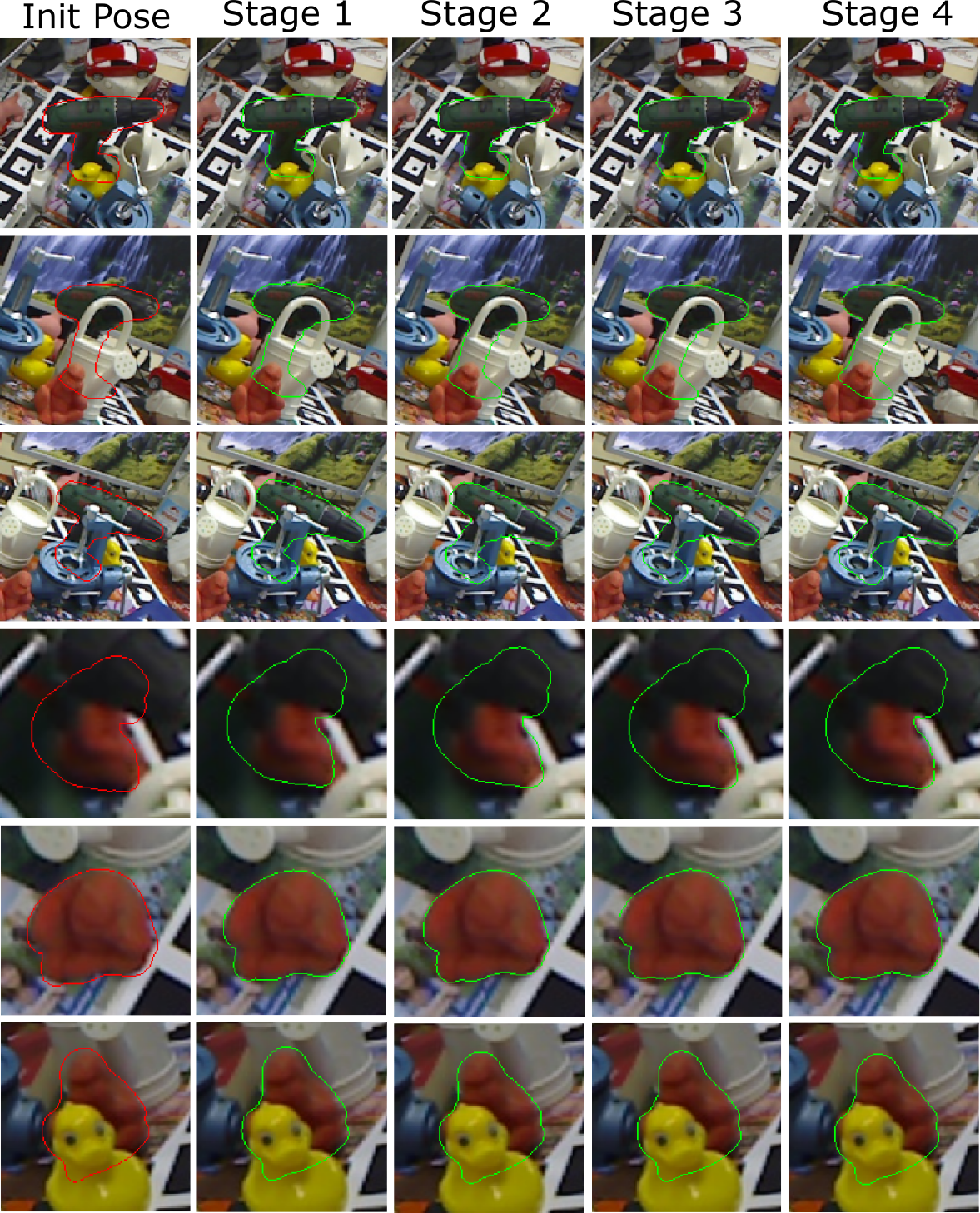}
    \caption{Iterative refinement results. For initialization,  we use PVNet \cite{peng2019pvnet}.  The red object outlines show the initial poses. The green outlines indicate the pose obtained via our refiner. We show the pose obtained after each step of the refiner. Notice that after each iteration, the rendered outline is closer to the real object outline.
}
    \label{fig:fig_5}
\end{figure*}

In this experiment, we provide a qualitative evaluation of our refiner. On a set of different examples, we show the initial 6D pose estimation and the predicted 6D pose after each stage of our model. We illustrate the poses in Fig.~\ref{fig:fig_4} and Fig.~\ref{fig:fig_5} by plotting the outline of the object at the predicted pose. We can notice that initial outlines are not matching the real object outlines. After each stage iteration, the outlines of the objects are closer and closer to the real object outlines. This experiment confirms the numerical results presented in the paper.

\end{document}